\documentclass[10pt,twocolumn,letterpaper]{article}

\usepackage[pagenumbers]{cvpr} 

\usepackage{graphicx}
\usepackage{amsmath}
\usepackage{amssymb}
\usepackage{booktabs}
\usepackage{multirow}
\usepackage{multicol}
\usepackage{float}
\usepackage[dvipsnames]{xcolor}
\usepackage[accsupp]{axessibility}  
\usepackage{booktabs}

\definecolor{slateblue}{RGB}{68, 119, 170}
\definecolor{cyan}{RGB}{102, 204, 238}
\definecolor{forestgreen}{RGB}{34, 136, 51}
\definecolor{brightgreen}{RGB}{26, 255, 26}
\definecolor{ochreyellow}{RGB}{204, 187, 68}
\definecolor{ultrared}{RGB}{238, 102, 119}
\definecolor{purple}{RGB}{170, 51, 119}
\definecolor{grey}{RGB}{187, 187, 187}

\usepackage[pagebackref,breaklinks,colorlinks]{hyperref}

\usepackage[capitalize]{cleveref}
\crefname{section}{Sec.}{Secs.}
\Crefname{section}{Section}{Sections}
\Crefname{table}{Table}{Tables}
\crefname{table}{Tab.}{Tabs.}

\def\cvprPaperID{13} 
\def\confName{EarthVision}
\def\confYear{2022}

\begin{document}

\title{Self-Supervised Learning to Guide Scientifically Relevant Categorization of Martian Terrain Images}

\author{
    Tejas Panambur$^{1,}$\thanks{Authors contributed equally. Correspondence to: {\tt \{tpanambur, dchakraborty\}@umass.edu}}
    \qquad Deep Chakraborty$^{2,}$\footnotemark[1]
    \qquad Melissa Meyer$^3$
    \qquad Ralph Milliken$^3$\\
    Erik Learned-Miller$^2$ 
    \qquad Mario Parente$^1$\\
    $^1$ Department of Electrical and Computer Engineering, UMass Amherst\\
    $^2$ Manning College of Information and Computer Sciences, UMass Amherst\\
    $^3$ Department of Earth, Environmental, and Planetary Sciences, Brown University
}
\maketitle

\begin{abstract}
  Automatic terrain recognition in Mars rover images is an important problem not just for navigation, but for scientists interested in studying rock types, and by extension, conditions of the ancient Martian paleoclimate and habitability. 
  Existing approaches to label Martian terrain either involve the use of non-expert annotators producing taxonomies of limited granularity (\eg soil, sand, bedrock, float rock, etc.), or rely on generic class discovery approaches that tend to produce perceptual classes such as rover parts and landscape, which are irrelevant to geologic analysis. Expert-labeled datasets containing granular geological/geomorphological terrain categories are rare or inaccessible to public, and sometimes require the extraction of relevant categorical information from complex annotations. In order to facilitate the creation of a dataset with detailed terrain categories, we present a self-supervised method that can cluster sedimentary textures in images captured from the Mast camera onboard the Curiosity rover (Mars Science Laboratory).
  We then present a qualitative analysis of these clusters and describe their geologic significance via the creation of a set of granular terrain categories. The precision and geologic validation of these automatically discovered clusters suggest that our methods are promising for the rapid classification of important geologic features and will therefore facilitate our long-term goal of producing a large, granular, and publicly available dataset for Mars terrain recognition. Code and datasets are available at \href{https://github.com/TejasPanambur/mastcam}{https://github.com/TejasPanambur/mastcam}.
\end{abstract}

\begin{figure*}[t]
    \centering
    \includegraphics[width=1.0\linewidth]{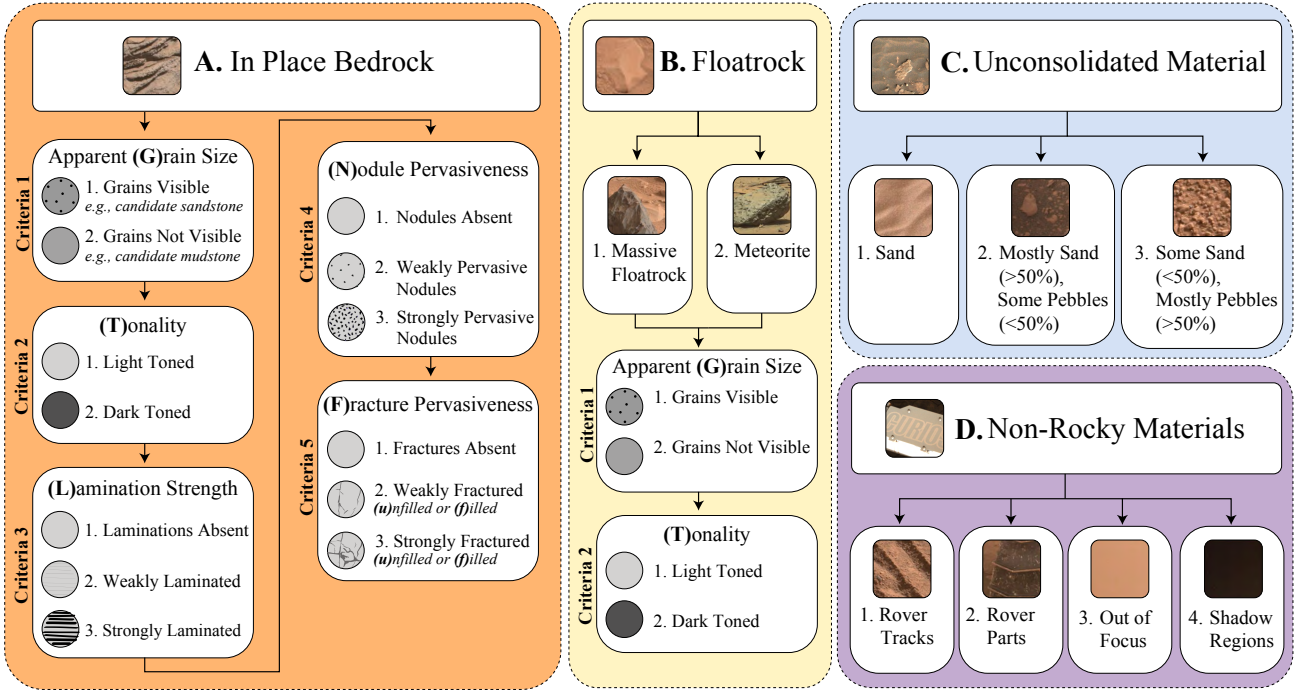}
    \caption{Taxonomy of geological classes (as found in our dataset). Note that a given class can be coded into any of the following categories using a combination of alphabets and numbers as shown in bold. For instance, weakly laminated red mudstone, which is non-nodular, and has calcium sulfate-filled hairline planar fractures can be represented as {\tt A-G2-T1-L2-N1-F2f}.}
    \label{fig:labels}
\end{figure*}


\section{Introduction}
\label{sec:intro}

Automatic terrain recognition has aided the navigational operations of Mars rovers by solving challenges such as traversability analysis \cite{spoc, ai4mars}, slip prediction \cite{gonzalez2018deepterramechanics, cunningham2017locally}, and minimizing driving energy \cite{higa2019vision}. A primary scientific objective of both the Mars Science Laboratory (MSL or Curiosity rover) \cite{msl} and Mars 2020 (Perseverance rover) \cite{mars2020} is to answer questions about water activity and the potential for past life on Mars. Toward this objective, a necessary step is to discriminate and/or correlate the various rock types and rock textures encountered along the rover traverse. The missions also seek to understand the geological history and evolution of the planet, and to prepare for future robotic and human exploration. These objectives are pursued through a plethora of geological and geochemical experiments onboard the rovers \cite{meyer2020microscale, mangold2017classification, cousin2017classification, cousin2019classification}. Crucially important are investigations related to imaging devices as they provide essential geologic context and highlight the presence of morphological or sedimentological features that may be indicative of water alteration (\eg the presence of fractures in-filled by veins or nodular structures that are directly responsible for chemical and mineralogical changes in the rock due to the interaction with water). The prospect of automating the difficult and somewhat subjective task of identifying and cataloging geomorphological and textural classes in Martian terrain would greatly improve the scientific return of Martian missions by allowing scientists to focus on more fundamental analyses, and speculations on formational mechanics, rather than classification. The need to provide scientists with relevant and granular terrain classes clearly arises.
Deep learning approaches have made great strides towards solving terrain classification tasks \cite{scoti, chakravarthy2021mrscatt, mslv1, mslv2, wang2021semi, gonzalez2018deepterramechanics, PALAFOX201748, spoc}, but the datasets and approaches available for rock classification are still very limited. 

In this paper, we develop an approach that can automatically assemble morphological and textural categories from the large amounts of unlabeled images available. An assessment from planetary scientists is used to validate our discoveries and their ability to represent clear scientific phenomena relevant to current scientific research.

Mars surface images originate from two primary sources: Mars rover missions \eg MSL, Mars2020, etc., \textit{and} Mars orbiter missions \eg Mars Global Surveyor (MGS), Mars Reconnaissance Orbiter (MRO), etc. While terrain can be identified both from ground and orbiter images, orbiter images are simply not acquired at scales fine enough to discern finer textural details that are necessary to uniquely identify categories related to rock types or the presence of diagnostic small-scale alterations. Therefore datasets constructed using these images \cite{ghosh2009automatic, spoc, domars16k} are not useful for the task at hand. Ground images captured from a variety of cameras mounted on rovers (\eg mastcam, navcam, and chemcam) on the other hand, are acquired at the necessary scale to facilitate detailed geomorphological analysis. However, efforts to label these images \cite{mslv1, mslv2, ai4mars} usually produce coarse labels such as ``sand", ``soil", and ``bedrock" as a result of non-expert annotations \textit{or} perceptual classes such as ``rover parts" and  ``landscape" as a result of generic class discovery approaches \cite{demud} precluding geologic analysis. Expert-labeled datasets on the other hand, exhibit more detailed terrain categories \cite{labelmars} but occasionally suffer from limited usability due to their complex annotation-based representations \cite{scoti}. Such labeled datasets are, furthermore, inaccessible to the public. 

This work is born from the desire to create a large, publicly accessible database of scientifically relevant textural/morphological terrain categories from readily available rover images, with a method that addresses the discrepancies described above. We propose a self-supervised deep clustering algorithm that can automatically group robust terrain categories by utilizing a network designed for texture recognition. A geology expert then provides a qualitative assessment of the discovered clusters and the scientific significance of such clusters. The expert also assigns a set of granular labels to images selected randomly from the discovered clusters. These labels have broad categories that can be further divided into subcategories based on different attributes (similar to LabelMars \cite{labelmars}). Here, top-level categories such as bedrock, floatrock, unconsolidated material, and non-rocky materials, have been further classified based on types of rock formations and other attributes such as grain size, tonality (hue), apparent lamination strength, nodule and fracture pervasiveness, as applicable (\Cref{fig:labels}). 
We don't claim that our work is a gold standard for terrain classification, but believe it to be a step in the right direction. Finally, we use the expert-derived labels to evaluate the quality of our clustering algorithm both by showing that it can produce homogeneous and well separated clusters, as well as evaluating its precision on a test set.

Our contributions in this paper are two-fold:\\
\begin{enumerate}
    \item We develop a novel synthesis of deep texture encoding techniques and self-supervised deep clustering algorithms to support rapid and robust terrain categorization.
    \item We produce an exhaustive taxonomy for classifying Martian terrain as seen in curiosity mastcam images, designed by a planetary geologist and supported with a review of how our approach could help geologic exploration.
\end{enumerate}

\section{Related work}
\label{sec:related}

\subsection{Efforts to label Martian terrain}
\label{subsec:mars-datasets}

Several efforts have been have been proposed for Mars terrain classification \cite{ai4mars, mslv1, mslv2, gmsri, labelmars, scoti, ghosh2009automatic, spoc, domars16k}. Earlier works, such as \cite{ghosh2009automatic, spoc, domars16k} annotated orbiter images from the MRO Context Camera (CTX) \cite{malin2007context} or High Resolution Imaging Science Experiment (HiRISE) \cite{hirise} to facilitate rover navigation. Although some of these works have defined important geomorphological categories, orbiter images with a resolution of around $25\ cm$/pixel are only partially suited for detailed geological analysis. This motivated the creation of labeled datasets using ground images captured from Mars rover cameras that have a much higher resolution ($150$ or $450\ \mu m$/pixel in the MSL right and left mastcam \cite{mastcam} for instance). It is the following works therefore, with which we compare and contrast in this paper. Wagstaff \etal curate a set of around $9000$ images in total using a combination of expert annotation and automatic class discovery using the DEMUD \cite{demud} algorithm \cite{mslv1, mslv2}. These labels include a variety of rover parts and artifacts, and a small set of geological categories such as float rock, layered rock, veins, sand, etc. AI4Mars \cite{ai4mars} is currently the largest labeled Mars terrain dataset, containing $326k$ images with semantic segmentation labels of categories such as soil, bedrock, sand, and big rock. These labels, while simple and useful for tasks such as navigation, are not granular enough for geological analyses. Inspired by the need for a content-based search system that enables scientists to interact with the rover using natural language descriptions, Qiu \etal propose SCOTI \cite{scoti}. They create a Mars image caption dataset starting from 1250 expert captioned images, training an image caption model, and progressively growing using predictions on unlabeled images with open/expert review. The final dataset contains more than $12,500$ images with natural language descriptions which include relevant geomorphological features together with general statements of limited relevance. A scientist seeking to carefully catalog the different textural/morphological categories in an image would find it difficult to extract a large set of uniquely identifiable class labels from the long descriptions. This annotated dataset is publicly unavailable. 
More similar to our effort is the the LabelMars project \cite{labelmars} that annotated 5000 images with the help of undergraduate geology students using a labeling scheme based on hierarchical morphological categories. Coarse categories such as ``sedimentary", ``magmatic", or ``meteoric" are further divided into sub-categories such as concretions/nodules and light/dark tonality, but this dataset is also publicly unavailable. Our proposed approach also discovers similar classes with high granularity that are labeled by an expert (see \cref{fig:labels}), albeit in an self-supervised fashion. This work pushes the labeling effort to the point that the identified categories reproduce exhaustively the set of criteria that are necessary for a complete characterization of the terrain that is possible by an expert with the imaging data available. Our dataset is described in detail in \cref{sec:dataset}.

\subsection{Terrain recognition}
\label{subsec:terrain}

Terrain recognition is a popular area of research in computer vision due to its vast applications in autonomous driving and terrain classification. It is usually cast as a texture recognition problem, and traditional approaches used geometric features, particularly curvature, color features, lighting, illumination direction, and photometric properties \cite{curve, color, lighting, visualapp}, followed by feature pooling as seen in bag-of-words models \cite{lazebnik2006beyond, csurka2004visual}. Recently, deep learning models with end-to-end texture/terrain recognition have gained popularity. However, naive CNN architectures are inadequate for texture recognition as they aren't invariant to spatial layout, and recognizing textures typically requires preserving some orderless information. Therefore, texture recognition approaches usually have distinct architectures compared to CNNs used for object recognition, to preserve fine textural details. Zhang \etal introduce a CNN architecture with end-to-end dictionary learning and feature pooling for orderless texture recognition \cite{depten}. Further, Xue \etal hypothesize that surfaces are not completely orderless, and propose the Deep Encoding Pooling Network (DEP) \cite{gtos-dep} that incorporates both orderless texture details and local spatial information, combined using bilinear pooling \cite{bilinear}. Several state-of-the-art approaches have since been proposed by integrating stronger geometric priors into deep networks or by encoding features from different layers in the network \cite{map-net, dsr-net, class-net, fe-net, multer}. However, our method is based off of Xue \etal \cite{gtos-dep} for its simplicity and adaptability to a self-supervised objective such as the one we use in this paper.

Self-supervised and unsupervised deep embedding learning is a relatively new and active area of research \cite{simclr, moco, swav, byol, simsiam, barlow, colorization, jigsaw, rotation}. A majority of the existing methods work fairly well on object-centric \cite{imagenet, cars, fgvc-aircraft, birds} or scene-centric \cite{voc, coco, lvis} datasets. However, very little research exists for unsupervised texture/terrain recognition \cite{tejas1, tejas2}. Panambur and Parente \cite{tejas1} use an alternating clustering and classification approach similar to DeepCluster \cite{deepcluster} to cluster mars terrain images. The resulting learned representations form homogeneous clusters of different kinds of terrain. In a followup work, Deep Clustering using Metric Learning (DCML) \cite{tejas2}, the authors train a triplet network by iteratively clustering the features and using the cluster labels to form triplets. The features generated have high \textit{inter}-class distance while preserving low \textit{intra}-class distance, making it ideal for retrieval tasks. However, there is no expert evaluation of the quality of clustering, unlike our work. Moreover, these approaches still suffer from the same drawbacks of traditional CNN architectures used for texture recognition. To rectify this problem, we incorporate the texture encoding module \cite{gtos-dep} into the CNN architecture to get better terrain recognition performance, while using the same metric learning approach  as \cite{tejas2} in order to leverage the large amounts of unlabeled Mars terrain images. We hypothesize that the learned representation can capture better the nuances between rocks that are geologically significant, while grouping similar types of terrain together.

\section{Deep self-supervised texture recognition}
\label{sec:texture} 
We first present the network architecture that is used in our approach in \cref{subsec:dep}. This architecture was specifically designed for texture classification, and modified to support self-supervised training. We then describe our self-supervised training objective in \cref{subsec:dcml}.

\subsection{Deep encoding pooling network}
\label{subsec:dep}
We start from the CNN architecture described in \cite{gtos-dep}. Given an input image $\mathbf{I}$, and the backbone feature extraction function $F_\theta$ (in this case ResNet-18 \cite{resnet}), we obtain features $\mathbf{X}_{f} = F_\theta(\mathbf{I})$ . The outputs from the feature extractor feed two separate layers in the network. One of these is a texture encoding layer \cite{depten} that produces an orderless representation of the features and preserves fine textural details, the kind that can be seen in terrain images. Given a feature map $\mathbf{X}_{f}$ and an encoding layer $F_{e}$ we obtain a texture embedding $\mathbf{X}_{t} = F_{e}(\mathbf{X}_f)$. 
The other layer is the usual global average pooling (GAP) layer that preserves spatial information, and the pooled embedding is defined as $\mathbf{X}_{g}$. 
The outputs of these layers are combined using a bilinear pooling layer \cite{bilinear} that helps in capturing the relationship between the orderless texture information and spatial information. The output embedding $\mathbf{X}_{b}$ from the bilinear layer combines texture features $\mathbf{X}_{t}$ and spatial features $\mathbf{X}_{g}$. Further, a fully connected layer and a linear classification layer are normally used to train the network using a supervised classification objective. We remove the final classification layer from the network that is generally used to map the embeddings to the fixed number of labeled categories in the dataset. Since we train our network on unlabeled data, we directly use the embeddings generated on the penultimate layer (fc-7 features) given by $\mathbf{X}_{emb}$.

\subsection{Deep clustering using metric learning}
\label{subsec:dcml}

The DCML algorithm alternatively clusters the embeddings using standard K-means clustering algorithm and uses
the subsequent assignments as pseudo-labels to train a metric learning objective\cite{tejas2}. We cluster the features from the embedding layer $\mathbf{X}_{emb}$, and assign pseudolabels to these clusters so that they can be used to generate triplets comprising an anchor, a positive, and a negative example. Positive examples are samples from the same cluster as the anchor, and negative examples belong to a different cluster. These triplets are used to minimize a distance metric objective called triplet loss. The triplet loss minimizes the distance between the anchor and positive examples and maximizes the distance between the anchor and negative examples in the embedded space. We use the same triplet loss objective with triplet sampling strategy defined in \cite{tejas2} as:
{\small
\begin{equation}
    \mathcal{L} = \sum_{i=1}^N ||\mathbf{X}_{{emb}_{a}}^i - \mathbf{X}_{{emb}_{p}}^i||_2^2 - ||\mathbf{X}_{{emb}_{a}}^i - \mathbf{X}_{{emb}_{n}}^i||_2^2 + \alpha
\end{equation}}

where $\mathbf{X}_{{emb}_{a}}^i$, $\mathbf{X}_{{emb}_{p}}^i$, $\mathbf{X}_{{emb}_{n}}^i$ are the embeddings of the anchor, positive, and negative examples, and $\alpha$ is the desired margin\cite{tejas2,triplet}. We set $\alpha=1$ in our experiments.

\section{Dataset}
\label{sec:dataset}

\subsection{Sourcing data and preprocessing}
\label{subsec:creation}
Our dataset consists of DRCL (decompressed, radiometrically calibrated, color corrected, and geometrically linearized) images \cite{malin2017mars} acquired by the MSL Mast cameras between sol 1 and sol 2800. We restrict our focus to terrain images by following the settings in \cite{tejas2} to eliminate the majority of images containing rover parts, sky, etc. In order to prevent scale disparities in terrain features, we limit our analysis to images in which the distance of a target to the rover is within $15m$. The resulting dataset contains $~30,000$ images of size typically $1200\times1600$ pixels. Since an image of this size might contain several different kinds of terrain, smaller patches of size $128\times128$ and $256\times256$ were extracted from it using a sliding window with a stride $50\%$ that of the patch size. The difference in patch size corresponds to the difference in focal lengths of the two cameras that make up the ``left eye" and ``right eye" of the curiosity rover, and were applied accordingly. This yields a total of $2.4M$ patches which is twice the size of the ImageNet-1k dataset \cite{imagenet}. 

A problem of class balancing still remained as a large number of patches contain mostly unconsolidated terrain and classes of geologic relevance are less frequent (long-tailed distribution). Therefore, in order to avoid any distribution mismatch between the training and test sets, we do not select patches randomly for each set (as is common practice). Instead, we follow a careful patch extraction strategy where patches in the training set are selected from the left $60\%$ portion of a given image, and patches from the remaining $40\%$ of the image are reserved for the test set. Criticism then arises that such an approach could lead to different views of the same geographical area appearing in both training and test sets due to panning of the camera. Though this is a rare occurrence (mastcam has a small field of view), two patches imaging the same area have different viewpoints and are as similar or different as two samples drawn from the same distribution albeit strongly correlated. Therefore, in addition to ensuring that the same ``patch" is never used for both training and testing, patches captured from the same \textit{site} and \textit{drive} of the rover are removed from the results before evaluation.

\subsection{Grouping data and discovering classes}
\label{subsec:annotation}
Once the model described in \cref{sec:texture} is trained on the unlabeled data obtained as above, we collect expert feedback on the model performance through a web-based user interface. We think that a natural way to do this is using query images to retrieve the top-$K$ most similar images from the dataset ($K$-Nearest Neighbors). The query images are randomly sampled from the clusters found by the model. The nearest neighbors are the top-$K$ images in the dataset whose embeddings (output of the penultimate fully connected layer of the trained network or fc-7 features) are closest (using a distance measure such as euclidean distance) to the embedding of the query image. We then display the query images along with top-$K$ nearest neighbors on a webpage which supports image annotation. The interface is shown in the appendix (\cref{fig:interface}). We ask an expert in planetary geology to do the following:
\begin{enumerate}
    \item Characterize and label the query image using detailed geological/geomorphological criteria.
    \item Assess how many of the top-$K$ neighbors belong to the same category as the query image and identify the mistakes.
    \item Evaluate the homogeneity of a randomly sampled subset of clusters produced by the network and comment on their geological relevance.
\end{enumerate}
Several geological categories naturally emerge from this process (See \cref{subsec:eval} for the analysis). The top level categories include ``bedrock", ``floatrock", ``unconsolidated material", and ``non-rocky materials". These are further subdivided based on rock type, grain size, tonality (hue), lamination strength, fracture and nodule pervasiveness, and preponderance of unconsolidated materials as apparent from the image. The full taxonomy is shown in \cref{fig:labels}.

We can traverse this hierarchy to generate detailed and uniquely-defined classes and also assign them a taxonomy code. For instance, a particular traversal could be represented as \verb|A-G2-T1-L2-N1-F2f|. Here \verb|A-G2| denotes bedrock with no visible grain (\eg mudstone), \verb|T1| is light-toned (in this case red colored), \verb|L2| indicates weakly laminated, \verb|N1| encodes the absence of nodules on this rock, and \verb|F2f| indicates that the rock is commonly (lightly) fractured with calcium sulfate filled veins. 
Note that since the above classifications were formulated by the expert after looking at the clusters produced by our model, our model reflects the kind of granular observations of terrain features a geologist would have to make as part of their scientific study (see discussion in \cref{subsec:eval}). Our taxonomy provides an exhaustive set of criteria that could be used to classify terrain, and leaves room for the addition of even finer categories still (which may not be present in our dataset), depending on the application, without any need to alter the hierarchy.

This level of granularity in categories in our dataset, while complex, is unprecedented for any terrain recognition dataset on Mars (and possibly Earth) and demands a high level of sophistication from automatic terrain classifiers. 
Note however that expert review is a long process, and due to the time available and also potential undersampling of certain categories, only a finite number of labels ($25$) could actually be identified in the sampled images. The full list of class descriptions along with their taxonomies is presented in the appendix (\cref{tab:classes}). Most of these classes are from the \textit{bedrock} category, as it is the focus of MSL missions and also a central category for geologic analysis. We plan to expand the number of classifications available for floatrocks and unconsolidated materials in future work.

\begin{figure*}[t]
    \centering
    \includegraphics[width=0.75\linewidth]{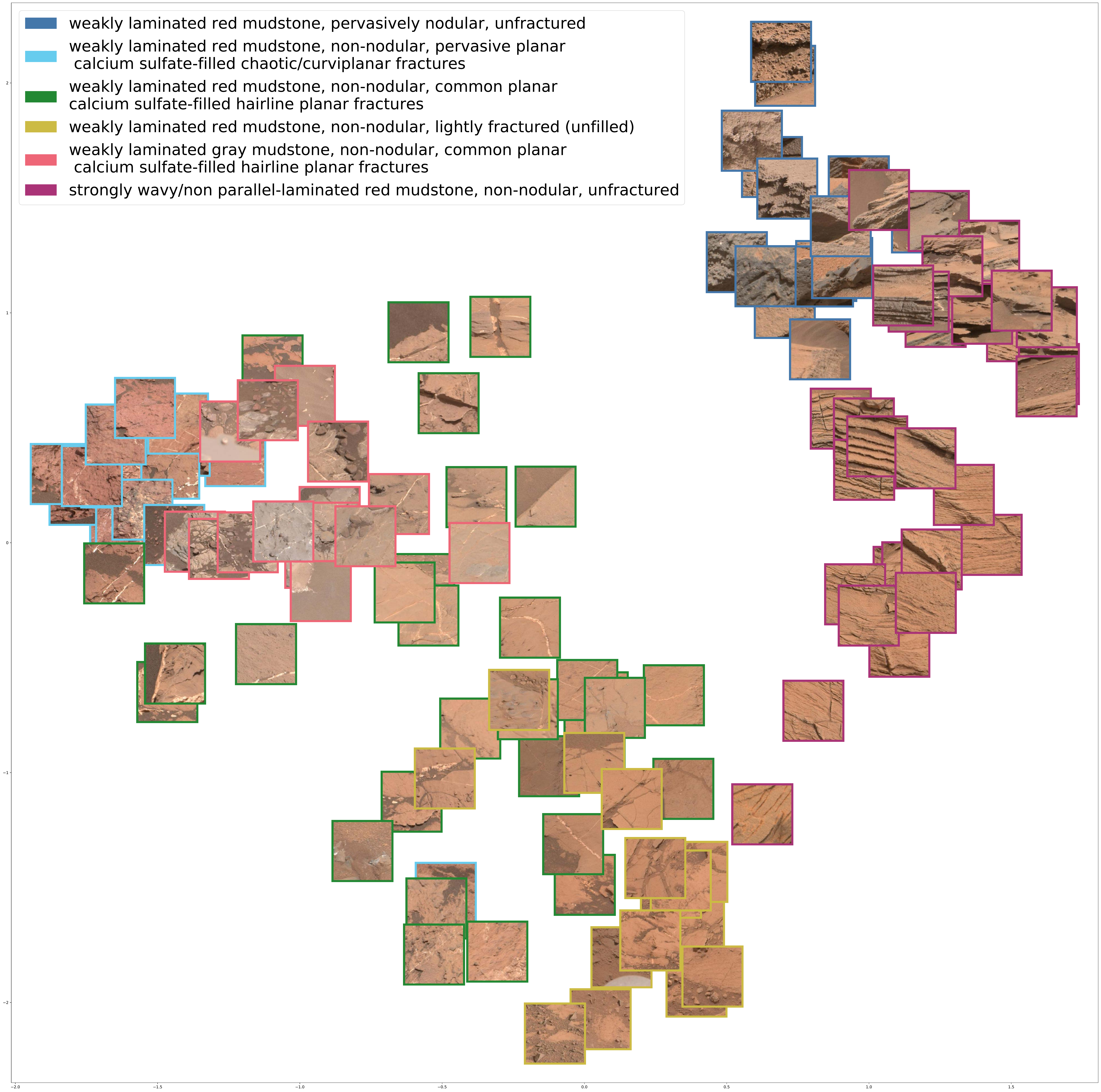}
    \caption{t-SNE\cite{tsne} visualization of 6 clusters from the model. The clusters show good separation, and overlap only when an increasing number of features are common (\eg \textcolor{forestgreen}{green} and \textcolor{ochreyellow}{yellow}). Best viewed in color (accessible) on a pdf processor with zoom.}
    \label{fig:tsne}
\end{figure*}
\begin{figure*}[t]
    \centering
    \resizebox{0.99\textwidth}{!}{
    \begin{tabular}{rl}
        Taxonomy &
        \begin{tabular}{p{0.05\linewidth}p{0.05\linewidth}p{0.05\linewidth}p{0.05\linewidth}p{0.05\linewidth}p{0.05\linewidth}p{0.05\linewidth}p{0.05\linewidth}p{0.05\linewidth}p{0.05\linewidth}p{0.05\linewidth}}
             Query & 1 & 2 & 3 & 4 & 5 & 6 & 7 & 8 & 9 & 10
        \end{tabular}\\
        {\small A-G1-T2-L3-N1-F1} & 
        \includegraphics[width=0.8\linewidth]{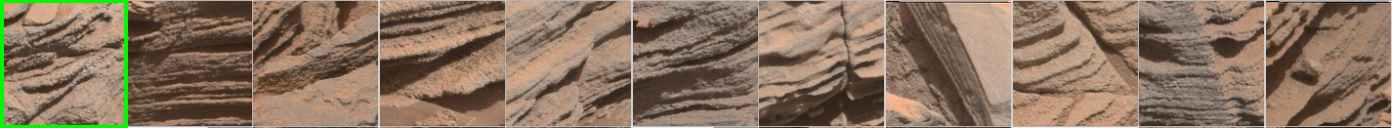} \\
        {\small C3} & 
        \includegraphics[width=0.8\linewidth]{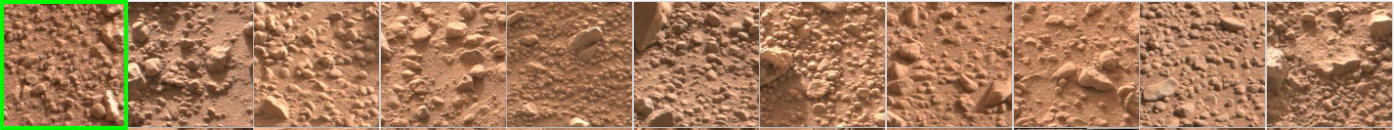} \\
        {\small A-G2-T1-L2-N1-F3f} & 
        \includegraphics[width=0.8\linewidth]{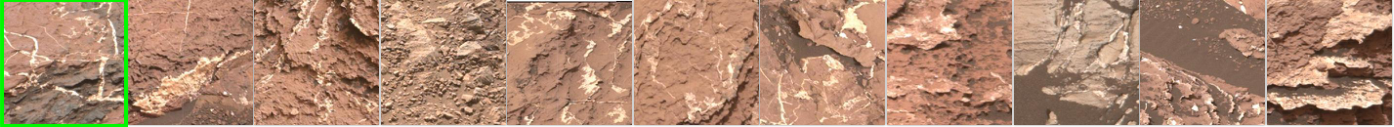} \\
        {\small A-G2-T2-L2-N1-F2f} & 
        \includegraphics[width=0.8\linewidth]{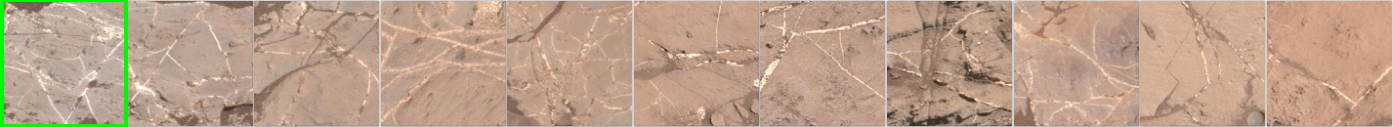} \\
        {\small A-G2-T1-L1-N1-F2u} & 
        \includegraphics[width=0.8\linewidth]{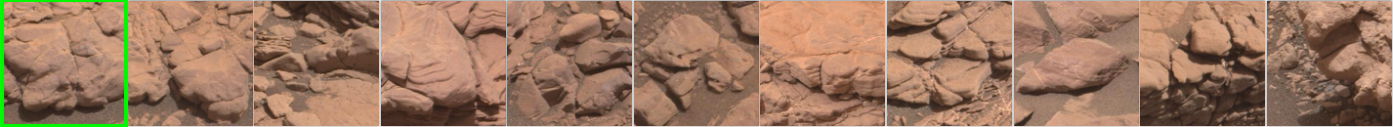} \\
        {\small A-G2-T1-L3-N1-F1} & 
        \includegraphics[width=0.8\linewidth]{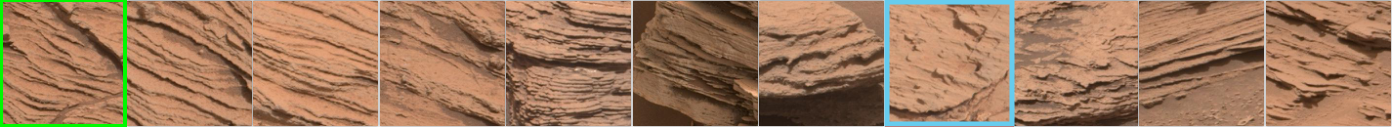} \\
        {\small C2} & 
        \includegraphics[width=0.8\linewidth]{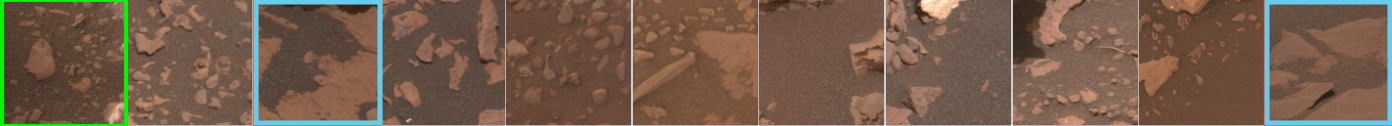} \\
        {\small A-G2-T1-L3-N1-F3u} & 
        \includegraphics[width=0.8\linewidth]{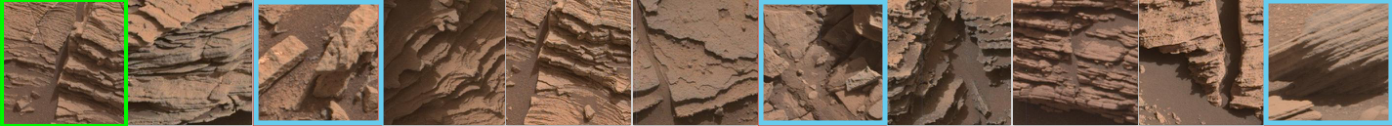} \\
        {\small A-G2-T1-L3-N1-F1} & 
        \includegraphics[width=0.8\linewidth]{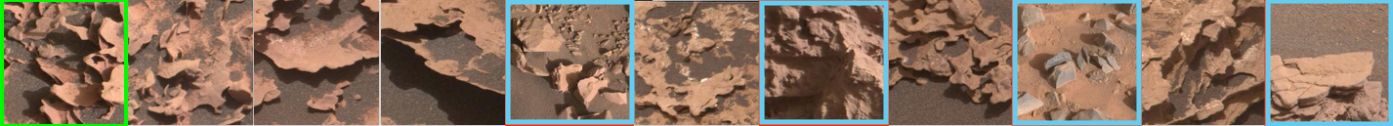} \\
        {\small A-G2-T1-L1-N2-F1} & 
        \includegraphics[width=0.8\linewidth]{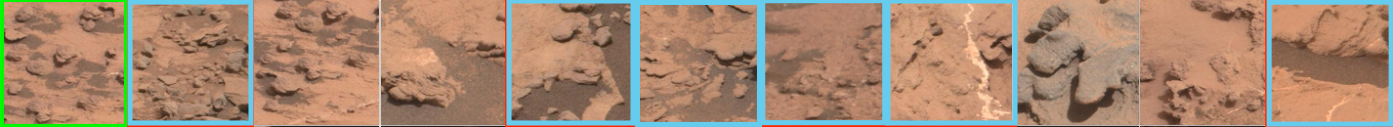} \\
    \end{tabular}}
    \caption{Top-$10$ nearest neighbors retrieved for $10$ query images (\textcolor{brightgreen}{green}). Nearest neighbors are from the test set, and restricted to be from different sites along the rover traverse. The rows are arranged in ascending order of mistakes (\textcolor{cyan}{cyan}) made by the model while retrieving patches. Only 10 out of 25 classes have been shown here along with their taxonomies. Best viewed in color (accessible).}
    \label{fig:knn}
\end{figure*}

\section{Experiments}
\label{sec:experiment}

\subsection{Implementation details}
\label{subsec:impl}

For the proposed method we use the DEP architecture \cite{gtos-dep} with ImageNet pretrained 18-layer ResNet\cite{resnet} as the backbone. 
The dimensionality of the embedding layer is set to $512$. 
For the clustering, the embeddings are PCA-reduced to 256 dimensions, whitened and $\ell^2$-normalized. 
We use Faiss K-means clustering algorithm \cite{faiss} with $K$ set to $150$ as determined by visual inspection of clusters. We use the SGD optimizer to train our network with a learning rate of $1\mathrm{e}{-4}$ and weight decay of $1\mathrm{e}{-5}$. The size of a minibatch corresponds to the number of samples per cluster times the number of clusters. We set number of samples per cluster as $4$, and the resulting batch size is $600$. We train on 4 Nvidia M40 GPUs, and training takes around 3 days. The criteria for convergence is the cluster stability over epoch. This is measured using Normalized Mutual Information (NMI) by calculating it between cluster assignments at epoch $t$ and $t-1$ \cite{deepcluster}. We find that the training saturates at epoch $40$ with NMI$=0.756$. We resize the image patches into $224\times224$. Since our model is trained to identify terrain features, it can be very sensitive to scale and orientation of the images. Therefore, we avoid using augmentations such as random resized crops and horizontal flips in order to avoid training on instances that might represent unrealistic viewpoints or non-existent geological formations.

\subsection{Evaluation}
\label{subsec:eval}

Our clustering performance is evaluated in two ways: using a visual inspection of homogeneity and cluster separation from t-SNE \cite{tsne} plots of learned embeddings, and computing the precision of a retrieval task from a test set given a query image. The precision of our model for retrieval tasks demonstrates the usefulness of our approach by allowing scientists to automate the process of finding visually similar terrain. Additionally, an expert opinion on the scientific significance of the clusters found by our model adds depth to the analysis of our model's performance and shows how our approach could be used to support geologic exploration.\\

\noindent \textbf{Qualitative.} \Cref{fig:tsne} shows a plot of 6 clusters obtained by projecting the embedding vectors for the points ($512$-dimensions) onto $2$-dimensions. Notice how the clusters are homogeneous and well separated in most cases, with occasional overlap in the case of highly similar terrain. We asked the expert to review a small subset of clusters produced by the model by looking at randomly sampled images from these clusters, and explain if our clusters could be useful for geological analysis the details of which follow.

Here, the clusters in \textcolor{forestgreen}{green} and \textcolor{ochreyellow}{yellow} represent two classes of weakly laminated red mudstone, that are non-nodular and fractured. The sole difference is that the fractures are either filled with calcium sulfate or unfilled, respectively. Using automation to make these sorts of subtle distinctions could be useful for scientists to rapidly make interpretations with important science and operational implications. In this example scenario, one could plausibly identify two generations of fracturing events within the dataset: one in which water was available to fill the fractures with calcium sulfate, and one in which it was not. Note how these clusters still have some overlap in the t-SNE plot which is consistent with their highly similar nature.

Another similar type of rock is seen in the \textcolor{ultrared}{red} cluster, where the rocks appear less red. This could either be because dust cover is minimized here, or because there is an inherent change in bedrock composition. This could be verified by spatially mapping these images and comparing whether these regions overlap with spectrally bland regions, which are often attributed to dust cover in CRISM data \cite{crism}. In yet another well separated cluster (\textcolor{purple}{purple}), we can see instances of well laminated, coherent bedrock from which sedimentary geologists typically infer ancient depositional processes. This cluster may be useful to this community as a way to rapidly identify exposure to facilitate more detailed sedimentological analysis.

Moreover, geologists are often interested in finding and mapping peculiarities in terrain, such as an increased amount of nodule formations in the bedrock as seen in the \textcolor{slateblue}{blue} cluster, or those which appear unusual such as the inordinately red rocks seen in the \textcolor{cyan}{cyan} cluster. The large number of images present make this task impossible to accomplish manually. Our model can quickly retrieve all images that are similar to a provided example, which could then be used to generate viewsheds on top of Mars orbital views \cite{arcgis} allowing geologists to corroborate their findings using global context. \\

\begin{table}[h]
\centering
\caption{Precision@$10$ obtained by expert review of each category present in our dataset. Only taxonomies are shown here to preserve space, the full description of classes is available in \protect\cref{tab:classes}}
\label{tab:precision}
\begin{tabular}{crc}
\hline
\textbf{ID} & \multicolumn{1}{c}{\textbf{Taxonomy}} & \textbf{Precision@10} \\\hline
1 & A-G1-T2-L3-N1-F1 & 1.0 \\
2 & A-G2-T1-L3-N1-F1 & 0.9 \\
3 & A-G2-T1-L3-N1-F1 & 0.9 \\
4 & A-G2-T1-L3-N1-F3u & 0.7 \\
5 & A-G2-T1-L3-N1-F1 & 0.6 \\
6 & A-G2-T1-L2-N1-F1 & 0.8 \\
7 & A-G2-T1-L2-N1-F2u & 1.0 \\
8 & A-G2-T1-L2-N1-F3u & 0.9 \\
9 & A-G2-T1-L2-N1-F2f & 1.0 \\
10 & A-G2-T2-L2-N1-F2f & 1.0 \\
11 & A-G2-T1-L2-N1-F3f & 1.0 \\
12 & A-G2-T1-L2-N3-F1 & 0.9 \\
13 & A-G2-T1-L2-N3-F3f & 1.0 \\
14 & A-G2-T1-L1-N1-F2 & 1.0 \\
15 & A-G2-T1-L1-N3-F1 & 1.0 \\
16 & A-G2-T1-L1-N2-F1 & 0.4 \\
17 & B1-G2-T1 & 0.7 \\
18 & B1-G2-T2 & 0.3 \\
19 & C1 & 1.0 \\
20 & C2 & 0.8 \\
21 & C3 & 1.0 \\
22 & D1 & 0.1 \\
23 & D2 & 0.9 \\
24 & D3 & 1.0 \\
25 & D4 & 1.0 \\\hline
\multicolumn{2}{r}{\textbf{Avg.}} & \textbf{0.836}\\\hline
\end{tabular}
\end{table}
\noindent \textbf{Quantitative.} Due to the lack of supervised annotations available for our data, we measure the performance of our model using a retrieval task. Given a query image $\mathbf I$ with known category $c_{\mathbf I}$, we poll the model to retrieve the top-$K$ images $\mathbf{I}_k'$ from the test set whose embeddings $F_\theta(\mathbf{I}_k')$ have the least distance to the embedding of the query image $F_\theta(\mathbf I)$. Here $F_\theta$ represents our deep network parameterized by $\theta$, and the distance measure selected is the euclidean distance between two vectors, $d(\mathbf u, \mathbf v) = ||\mathbf u - \mathbf v||_2$. We use the metric Precision@$K$ to evaluate the quality of the retrieved images. Precision@$K$ is defined as follows:
\begin{equation}
    \text{Precision@}K \triangleq \frac{1}{K}\sum_{k \in {1..K}} \mathbb{I}(c_{\mathbf{I}} = c_{\mathbf{I_k'}})
\end{equation}
where $\mathbb{I}$ is the indicator function. Here, we use $K=10$. Since we don't have actual labels available for the retrieved images, we once again seek expert review to correctly identify the classes of all retrieved images. The results for each category are shown in \cref{tab:precision}. \Cref{fig:knn} shows a subset of labeled query images from our dataset and the retrieved nearest neighbors. The nearest neighbors are restricted to be from different sites along the rover traverse in the Gale crater. This demonstrates our model's generalization performance and ability to retrieve interesting terrain images from different locations using a query image. This mechanism could potentially be used to propagate the label of a query image to similar \textit{unlabeled} images and overlaid on a map to study terrain changes along the rover traverse. Our model obtains an overall Precision@$10$ of $83.6\%$. Queries from 12 (out of 25) classes in our data have $100\%$ precision of retrieval (subset shown in first 5 rows in \cref{fig:knn}) and the performance degrades slightly for other classes. Common failure modes include bedrock with nodules confused with pebbles on the surface of the bedrock (Row 10 in \cref{fig:knn}) and rover tracks mistaken for fractured or strongly laminated bedrock (not shown). The former is a result of natural scale variations in the dataset due to the focus distance and the model's inability to capture them, whereas the latter happens in part because of the high visual similarity of sand tracks to laminated rocks and in part because of limited data available for such classes.

\section{Conclusion}
\label{sec:conclusion}
We presented a framework that would allow the creation of a large database of robust, readily available, and geologically relevant terrain categories of the Martian surface based on mastcam images. Our self-supervised deep clustering algorithm can automatically identify nuanced terrain categories by utilizing a network designed for texture recognition. The automatically discovered clusters enabled the creation of a robust taxonomy of scientifically relevant terrain categories through expert assessment, that can be used to rapidly label images. The granularity and homogeneity of the discovered clusters were evaluated using such labels qualitatively and quantitatively. 
The agreement between the membership identity of the clusters and the expert terrain categories show promise for extensive, automated analysis of geologic features on the Martian surface pertaining to a better understanding of depositional processes and interpreting the paleoclimate to ultimately answer the question of whether life once existed on Mars.

\section*{Acknowledgements}

This work was sponsored in part by the AFRL and DARPA under agreement FA8750-18-2-0126, and by NASA under grant 80NSSC19K1227. Part of this work was conducted on HPC equipment managed by the MassTech Collaborative. The authors would like to thank the anonymous reviewers for their invaluable comments, and Debasmita Ghose, Akanksha Atrey, and Karan Rakesh for proofreading the manuscript. 

{\small
\bibliographystyle{ieee_fullname}
\bibliography{ms}
}

\clearpage
\onecolumn
\appendix

\renewcommand\thefigure{\thesection.\arabic{figure}}
\renewcommand{\thetable}{\thesection.\arabic{table}}
\section{Appendix}
\setcounter{figure}{0}
\setcounter{table}{0}
\begin{table*}[h]
\centering
\caption{Full description of classes found in our database along with their taxonomy.}
\label{tab:classes}
\resizebox{0.88\textwidth}{!}{
\begin{tabular}{crl}
\hline
\textbf{ID} & \multicolumn{1}{c}{\textbf{Taxonomy}} & \textbf{Class Description} \\\hline
1 & A-G1-T2-L3-N1-F1 & strongly laminated dark-colored sandstone (grains visible), non-nodular, unfractured \\
2 & A-G2-T1-L3-N1-F1 & strongly wavy/non parallel-laminated red mudstone, non-nodular, unfractured \\
3 & A-G2-T1-L3-N1-F1 & strongly even/parallel-laminated red mudstone, non-nodular, unfractured \\
4 & A-G2-T1-L3-N1-F3u & strongly laminated red mudstone, non-nodular, pervasively fractured (unfilled) \\
5 & A-G2-T1-L3-N1-F1 & strongly laminated red mudstone, non-nodular, unfractured, platy weathered \\
6 & A-G2-T1-L2-N1-F1 & weakly laminated red mudstone, non-nodular, unfractured \\
7 & A-G2-T1-L2-N1-F2u & weakly laminated red mudstone, non-nodular, lightly fractured (unfilled) \\
8 & A-G2-T1-L2-N1-F3u & weakly laminated red mudstone, non-nodular, heavily fractured (unfilled) \\
9 & A-G2-T1-L2-N1-F2f & weakly laminated red mudstone, non-nodular, common planar calcium sulfate-filled hairline planar fractures \\
10 & A-G2-T2-L2-N1-F2f & weakly laminated gray mudstone, non-nodular, common planar calcium sulfate-filled hairline planar fractures \\
11 & A-G2-T1-L2-N1-F3f & weakly laminated red mudstone, non-nodular, pervasive planar calcium sulfate-filled chaotic/curviplanar fractures \\
12 & A-G2-T1-L2-N3-F1 & weakly laminated red mudstone, pervasively nodular, unfractured \\
13 & A-G2-T1-L2-N3-F3f & weakly laminated red mudstone, pervasively nodular, abundant calcium-sulfate filled hairine fractures \\
14 & A-G2-T1-L1-N1-F2 & apparently unlaminated /massive red mudstone, non-nodular, commonly fractured (unfilled) \\
15 & A-G2-T1-L1-N3-F1 & apparently unlaminated red mudstone, pervasively nodular, unfractured \\
16 & A-G2-T1-L1-N2-F1 & apparently unlaminated red mudstone, common protruding raised nodules, unfractured \\
17 & B1-G2-T1 & Red/Light-toned, massive floatrock \\
18 & B1-G2-T2 & Dark-toned, massive floatrock \\
19 & C1 & Sand \\
20 & C2 & Mostly Sand ($>50\%$), Some Rounded Pebbles ($<50\%$) \\
21 & C3 & Some Sand ($<50\%$), Mostly Rounded Pebbles ($>50\%$) \\
22 & D1 & Rover Sand Tracks \\
23 & D2 & Rover Parts \\
24 & D3 & Out of Focus \\
25 & D4 & Heavily Shadowed Areas\\\hline
\end{tabular}}
\end{table*}
\begin{figure*}[h]
    \centering
    \includegraphics[width=0.88\linewidth]{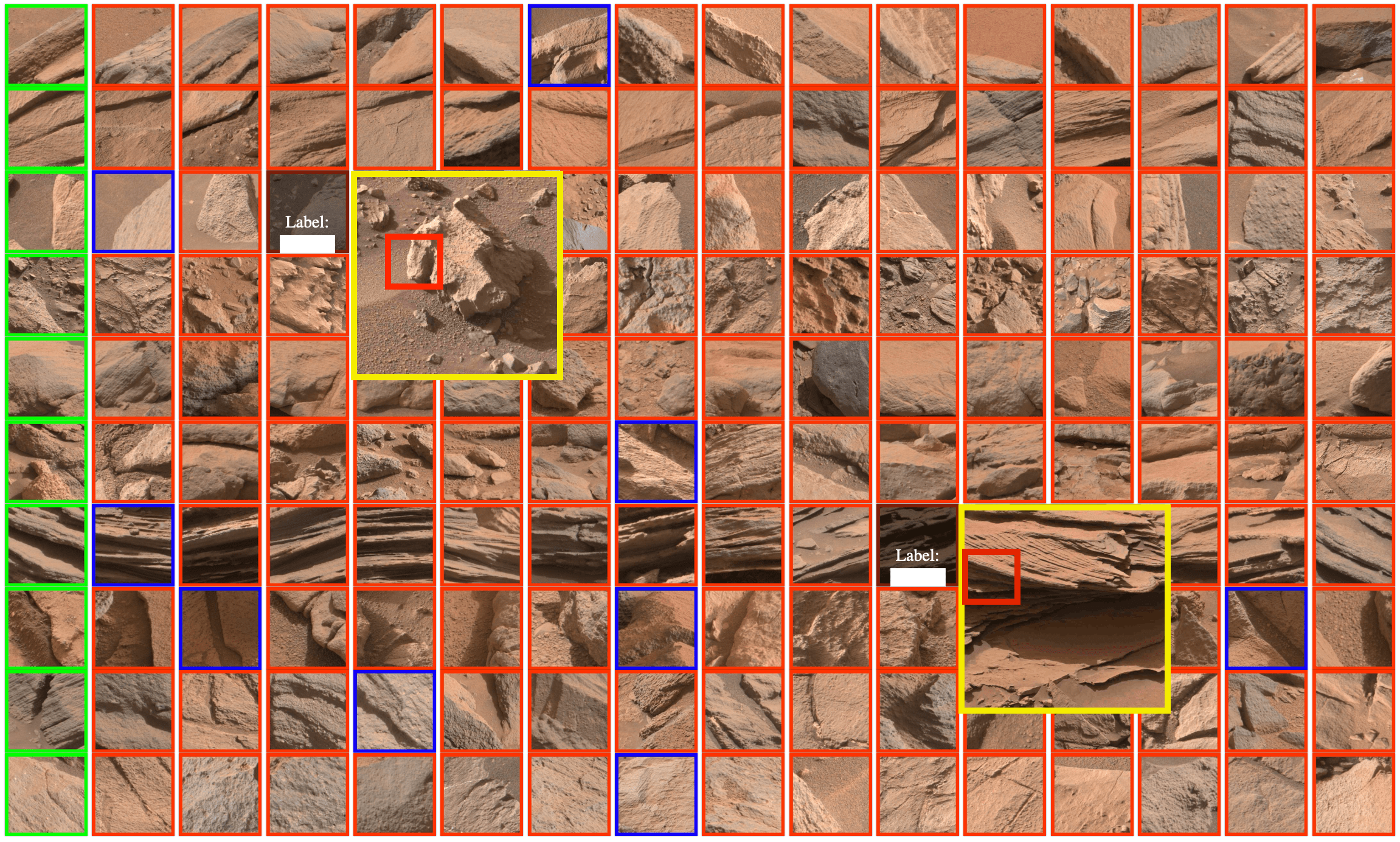}
    \caption{Labeling interface with query patches in the first column (green), and top-$K$ nearest neighbors ($K=15$) in the corresponding rows (red/blue indicates that image was captured from a different/same site). Left click allows user to enter a label as shown in (row $3$, col $4$), (row $7$, col $11$). Right click allows user to view the original image (yellow boxes) that the patch was drawn from (patch localized in red box) to verify if it is bedrock/floatrock/unconsolidated. Highlighted in this figure are instances of light-toned massive floatrock (row $3$, col $5$) and laminated red mudstone bedrock (row $7$, col $12$). Figure is best viewed in color.}
    \label{fig:interface}
\end{figure*}


\end{document}